\pdfoutput=1

\documentclass[11pt]{article}

\usepackage[preprint]{acl}

\usepackage{times}
\usepackage{latexsym}

\usepackage[T1]{fontenc}

\usepackage[utf8]{inputenc}

\usepackage{microtype}

\usepackage{inconsolata}

\usepackage{graphicx}

\usepackage{multirow}

\title{Capacity Matters: a Proof-of-Concept\\for Transformer Memorization on Real-World Data 
\thanks{This work has been accepted for publication at the First Workshop on Large Language Model Memorization (L2M2) at ACL 2025, Vienna, Austria.}
}

\author{Anton Changalidis \and Aki Härmä \\ \\
        Department of Advanced Computing Sciences (DACS),\\Faculty of Science and Engineering,\\Maastricht University, The Netherlands\\
        \texttt{anton@bioinf.me}, \texttt{aki.harma@maastrichtuniversity.nl}
        }

\begin{document}
\maketitle
\begin{abstract}
This paper studies how the model architecture and data configurations influence the empirical memorization capacity of generative transformers. The models are trained using synthetic text datasets derived from the Systematized Nomenclature of Medicine (SNOMED) knowledge graph: triplets, representing static connections, and sequences, simulating complex relation patterns. The results show that embedding size is the primary determinant of learning speed and capacity, while additional layers provide limited benefits and may hinder performance on simpler datasets. Activation functions play a crucial role, and Softmax demonstrates greater stability and capacity. Furthermore, increasing the complexity of the data set seems to improve the final memorization. These insights improve our understanding of transformer memory mechanisms and provide a framework for optimizing model design with structured real-world data.


\end{abstract}


\section{Introduction}
Transformer-based Large Language Models (LLMs) have revolutionized natural language processing, excelling at tasks ranging from text generation and translation to question answering and summarization. Despite these advances, a fundamental understanding of how these models store and recall information, particularly factual or structured knowledge, remains limited. Clarifying these mechanisms is crucial for optimizing model performance and enabling efficient, real-world deployment. One impactful example is healthcare, where transformer-based models could assist clinicians through wearable devices such as smart glasses or watches \citep{gupta_comprehensive_2024,wu_surgbox_2024,balloccu_ask_2024}. Due to privacy and reliability, the preferred system would be a local on-edge, requiring minimal computation but with the capacity to memorize all relevant facts in the specific healthcare area. 

Recent theoretical and empirical studies have sought to quantify the memorization capacity of transformers. \citet{kim2023provable} introduced mathematical bounds for memory capacity, demonstrating that transformers could memorize $O(d + n + \sqrt{nN})$ parameters, where $d, n, N$ correspond to embedding dimensions, dataset size, and model size, respectively. Additionally, \citet{kajitsuka2024optimalmemorizationcapacitytransformers} proved, that $\tilde{O}(\sqrt{nN})$ parameters are not only sufficient, but also necessary for some types of transformers. \citet{mahdavi2024memorizationcapacitymultiheadattention} extended this work by analyzing the effects of multi-head attention on memorization, revealing the interplay between architectural components and the model's ability to store and recall information. The experiments in \citet{härmä2024empiricalcapacitymodelselfattention} used randomly generated sequences of numbers to evaluate the memorization capabilities of the transformer models on unstructured data. Most capacity studies use synthetic datasets because accurate capacity measurement becomes very difficult in the case of uncontrolled free text content. 

The experiments reported in the current paper use sequential data generated from the knowledge graph, which, while controlled, has some of the hierarchical and relational complexity of real-world text content. More specifically, small-scale decoder-only transformer models \citep{brown2020languagemodelsfewshotlearners} were trained to memorize structured sentences derived from the Systematized Nomenclature of Medicine (SNOMED) knowledge graph (KG) \citep{El-Sappagh2018}, a comprehensive medical ontology, which encodes semantic relationships between medical concepts, offering a rich dataset to explore memory mechanisms under realistic conditions. Exact memorization of selected relations would be critical, for example, in the healthcare use cases described above. Our aim is not to generalize to all LLMs or domains, but rather to offer a practical, reproducible framework for measuring memorization on realistic KG data. The relative task simplicity is by design: more complex or less-controlled tasks would conflate memorization with generalization, making it difficult to draw clear, interpretable conclusions about model capacity.

To measure the memorization of the transformer models, the Maximum Attainable Capacity (MAC) method was used. It evaluates the practical limit of samples a model can retain when trained on a large dataset. Our approach leverages structured datasets consisting of static triplets and longer sequences simulating graph traversal paths, capturing relationship patterns between concepts. These datasets allowed us to empirically analyze how model architecture, training configurations, dataset size, and complexity influence training dynamics and final memorization performance.

This work serves as a proof-of-concept, showing that structured data in the real world can evaluate memorization in practice. Firstly, we introduce a reproducible pipeline for converting large ontologies into tokenized datasets suitable for memorization studies. Secondly, we evaluate how transformers' architecture influences capacity, building on prior theoretical insights. Lastly, we highlight cases where models fail to memorize all samples despite sufficient capacity, motivating future studies into training dynamics and error patterns.

Our findings do not aim to establish universal scaling laws or generalization behavior but to provide a reproducible framework for studying memory-limited models under realistic constraints.

\section{Methods}

\subsection{Data}

\subsubsection{Data Source and Preprocessing}

To evaluate transformer memorization and retrieval capabilities, we used SNOMED KG, which encodes medical concepts and their relationships as nodes and edges of a graph. It was accessed using the \texttt{owlready2} library \citep{lamy2017owlready}, filtering out non-informative or overly specific properties to ensure meaningful relationships. Unlike graph transformers that use GNNs \citep{shehzad2024graphtransformerssurvey}, we focus on a universal architecture, transforming the graph into (1) triplets (concept-property relationships, see \ref{tripgen}), and (2) sequences, simulating graph traversal paths (see \ref{sec:2.1.3}).

\subsubsection{Triplets Generation}
\label{tripgen}

\begin{figure}[hbt!]
    \centering
    \includegraphics[width=0.48\textwidth]{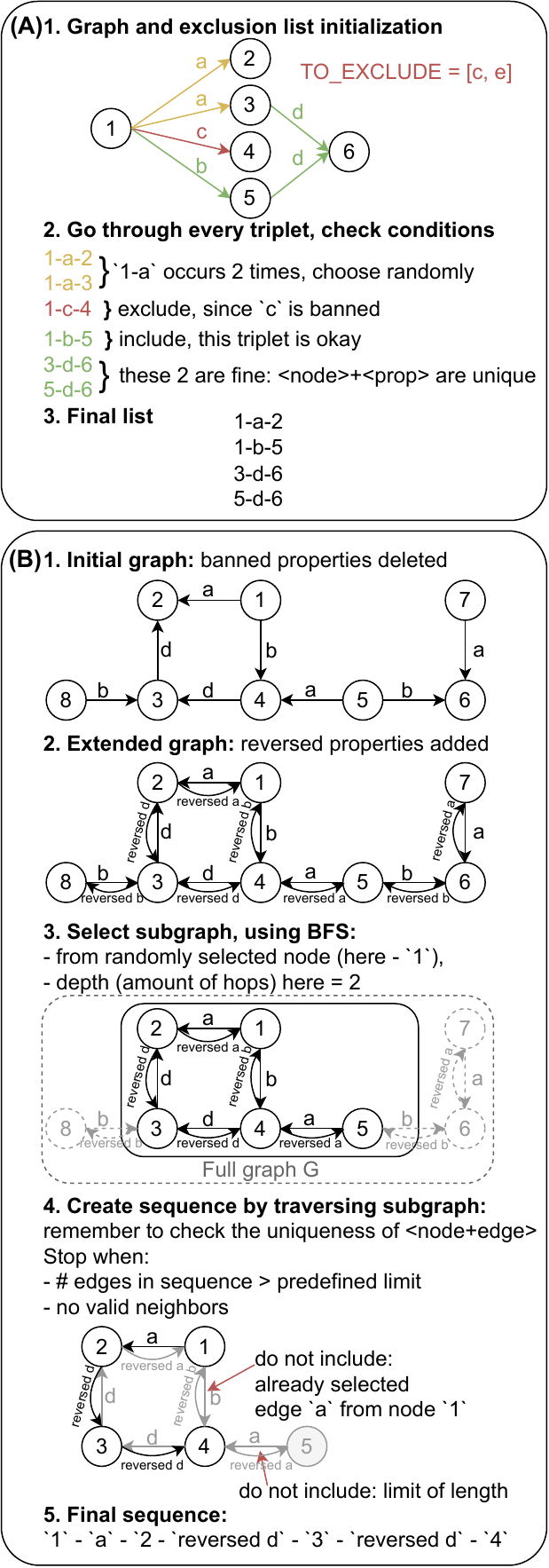}
    \caption{Algorithms of triplets (A) and sequences (B) data generation.}
    \label{fig1}
\end{figure}

A dataset of the form \texttt{(Concept, Property, Related Concept)} was created, capturing semantic relationships in the SNOMED KG (see Figure \ref{fig1}A). It involves graph initialization and the exclusion of non-informative properties, followed by the triplets extraction: for each concept in the KG, all allowed properties and their associated related concepts are retrieved. If multiple related concepts existed for a \texttt{(Concept, Property)} pair, one was randomly chosen to ensure uniqueness.

\subsubsection{Sequences Generation}
\label{sec:2.1.3}

The sequence generation simulated graph traversal to encode both local and global structures (Figure \ref{fig1}B). The extended graph excluded banned properties and added reverse edges for bidirectional traversal; labels were standardized. Sequences of the form \texttt{(node$_1$, edge$_1$, node$_2$, …, node$_{n-1}$, edge$_{n-1}$, node$_{n}$)} were generated by selecting a random starting node, creating a subgraph by breadth-first search (BFS) with a set depth and randomly traversing unique edges. Every time, check that the same \texttt{(node, edge)} pair is not already visited before. The traversal stopped once it reached a pre-defined random edge limit or when no valid neighbors remained. This process was repeated for the desired number of sequences.

\subsection{Transformers training}

Decoder-only transformers with variations in architecture were implemented. Each unique element (node or edge) was assigned a unique integer (ensuring that repeated elements were consistently tokenized), followed by learned positional encoding. The architecture included an embedding layer to map tokenized inputs into continuous vector representations, transformer decoder layers with multi-head attention mechanisms, and a linear output for token prediction. 

For all experiments, the task was to predict a concept based on the previous concepts and relations. The accuracy was evaluated as: $\frac{\#correct\_predictions}{\#total\_predictions}$  – the proportion of correctly predicted related concepts to the total number of predictions. Additionally, Maximum Attainable Capacity (MAC) was used as a more suitable metric to measure the capacity of the model. MAC is a computationally efficient alternative to the Maximum Library Size (MLS) method. While MLS involves iteratively training models on progressively larger datasets to determine the largest library size that can be fully memorized, MAC is measuring the maximum number of samples that a model can memorize, provided with a large library. Previous research has shown a strong correlation between MLS and MAC \citep{härmä2024empiricalcapacitymodelselfattention}, 
making MAC an effective and time-efficient choice for this study.

To minimize the effect of randomness, each experiment was repeated $10$ times for the first two setups and $3$ times for the third and fourth setups, reporting the mean and double standard deviation. Training accuracy was evaluated at every other epoch for all configurations.

Models were implemented in \texttt{PyTorch v1.13.1+cu117} \citep{paszke2017automatic} and \texttt{Transformers v4.30.2} \citep{DBLP:journals/corr/abs-1910-03771}, trained with cross-entropy loss and Adam optimizer (learning rate $0.001$) \citep{kingma2017adammethodstochasticoptimization}. All other were default unless specified. In total, $546$ models were trainded on NVIDIA A100 GPU with 16GB memory, totaling approximately $3,100$ hours of training time. Model sizes ranged from $2.9$ to $44.5$ million parameters, primarily varying with embedding size and layer count, but also influenced by vocabulary size.

\subsection{Code availability}

All code pertinent to the methods and results presented in this work is available at: \url{https://github.com/um-dacs-nlp/capacity/}.

\subsubsection{Triplets memorization}
\label{sec:2.2.1}

Three experimental setups were designed for the triplets dataset. In all cases, the prediction of a related concept was based on a unique concept-relation pair, making correctness unambiguous. 

In the first setup, dataset sizes ranged $50{,}000$ to $100{,}000$ samples. The model architecture consisted of a single transformer layer (embedding size $128$, 4 attention heads, Rectified Linear Unit (ReLU) activation function \citep{agarap2019deeplearningusingrectified}, batch size $64$, 500 epochs). This setup focused on evaluating memorization performance under a fixed architecture while varying dataset sizes. 

The second setup varied both architecture and activations:  transformer layers ($1$, $2$, or $4$), and activation functions (ReLU, Gaussian Error Linear Unit (GELU) \citep{hendrycks2023gaussianerrorlinearunits}, Randomized Leaky Rectified Linear Unit (RReLU) \citep{xu2015empiricalevaluationrectifiedactivations}, and Softmax \citep{boltzmann1868studien}), with dataset sizes of $50{,}000$, $70{,}000$, or $100{,}000$. To ensure fair comparisons, the total number of model parameters was kept constant across configurations by adjusting the embedding size (\texttt{d\_model} parameter in PyTorch implementation of Transformers) proportionally to the number of layers, using the formula: $\texttt{embedding\_size}=\Bigl \lfloor \frac{\texttt{base\_number\_of\_parameters}}{\texttt{n\_layers}}\Bigr \rfloor$  with a base number of parameters of $128$. This approach ensured that variations in performance could be attributed solely to architectural differences rather than changes in the total parameter count. For this setup, however, the batch size was increased to $128$ and models were trained for $1000$ epochs, since it was required for achieving a plateau.

The third setup examined the interplay between model depth and embedding size, while keeping other hyperparameters the same: number of layers was set to $1$ or $2$ and base numbers of parameters for embedding sizes varied in $\{16; 32; 64; 128\}$ (calculated as in the second experiment), with dataset sizes of $1{,}000$, $10{,}000$, $50{,}000$, and $100{,}000$. Only the Softmax activation function and $4$ attention heads were used. To ensure fair comparisons, the configurations were designed to evaluate the impact of increasing the embedding sizes and depth of the model on the performance of the memory. The total parameter count was recalculated for each configuration using the same formula as in the second experiment. For this setup, the batch size was $128$ and the training lasted $500$ epochs.

\subsubsection{Sequences memorization}

\begin{figure*}[ht]
    \centering
    \includegraphics[width=0.9\textwidth]{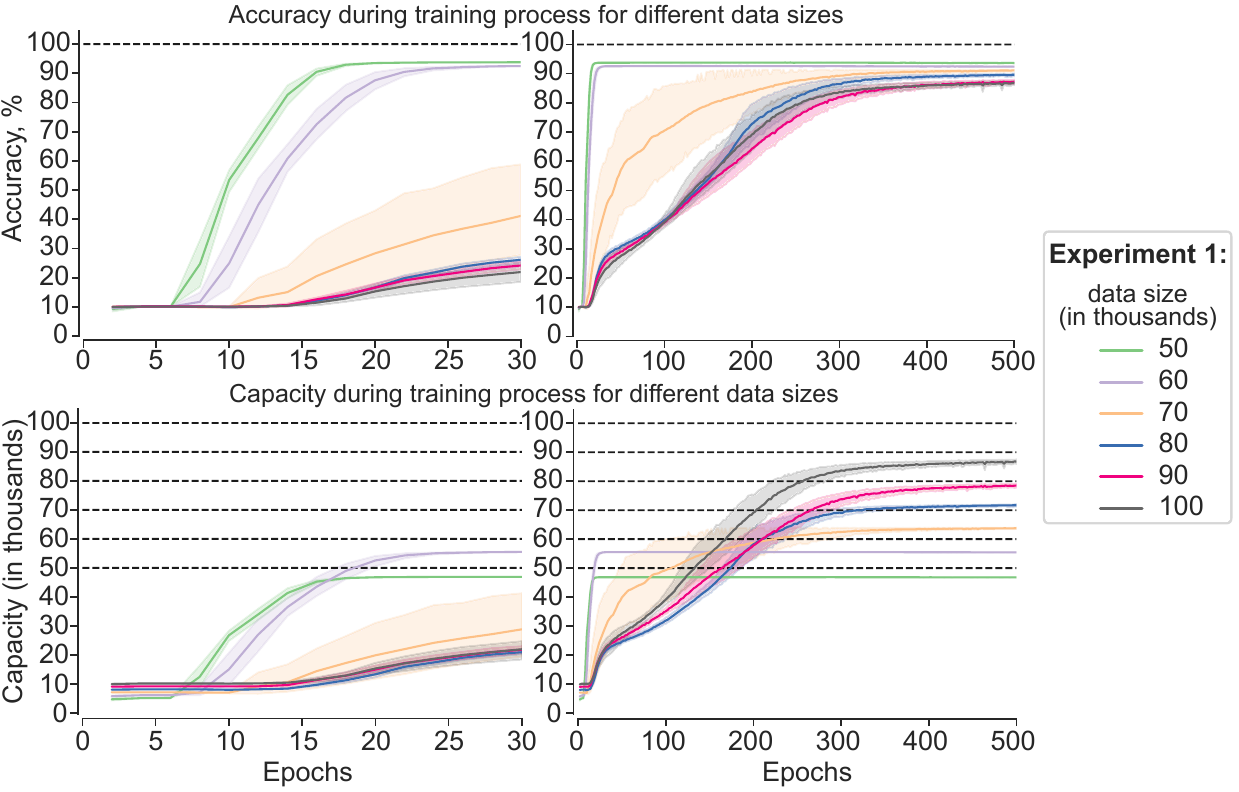}
    \caption{Trends in training accuracy (upper) and capacity (lower) for the first setup (different data sizes, for triplets dataset). Left: first $30$ epochs; right: full training process of 500 epochs.}
    \label{fig2}
\end{figure*}

The sequence memorization dataset used the same tokenization process as triplets, with additional steps for standardization: zero-padding at the end to a uniform length served both as a filler and a marker for sequence termination. A node mask was applied to distinguish the node from edge tokens for metric computation. Notably, each node was predicted based on all preceding tokens in the sequence, meaning the last node in a sequence benefited from the most context. This setup provided deeper insights into the transformer model's ability to handle more structured data and its patterns.

The experimental setup was consistent with the triplet setups: embedding size $64$, 4 attention heads, batch size $128$, and $400$ training epochs. Models with $1$, $2$, or $4$ layers were tested, using RReLU and Softmax activations. Dataset sizes were $20{,}000$, $50{,}000$, and $100{,}000$ sequences, each containing $4$–$6$ nodes ($3$–$5$ edges), built from subgraphs extracted via BFS with a depth of $5$ hops.

For this experiment, accuracy and capacity were measured similarly to the triplet-based experiments, with slight adaptations to account for the sequential structure of the data. Accuracy was defined as the proportion of correctly predicted tokens at node positions to the total number of node predictions in the dataset and is equal to all nodes across all sequences, excluding starting points. The total correct predictions also represent the MAC.

\section{Results}

\begin{figure*}[ht]
    \centering
    \includegraphics[width=0.99\textwidth]{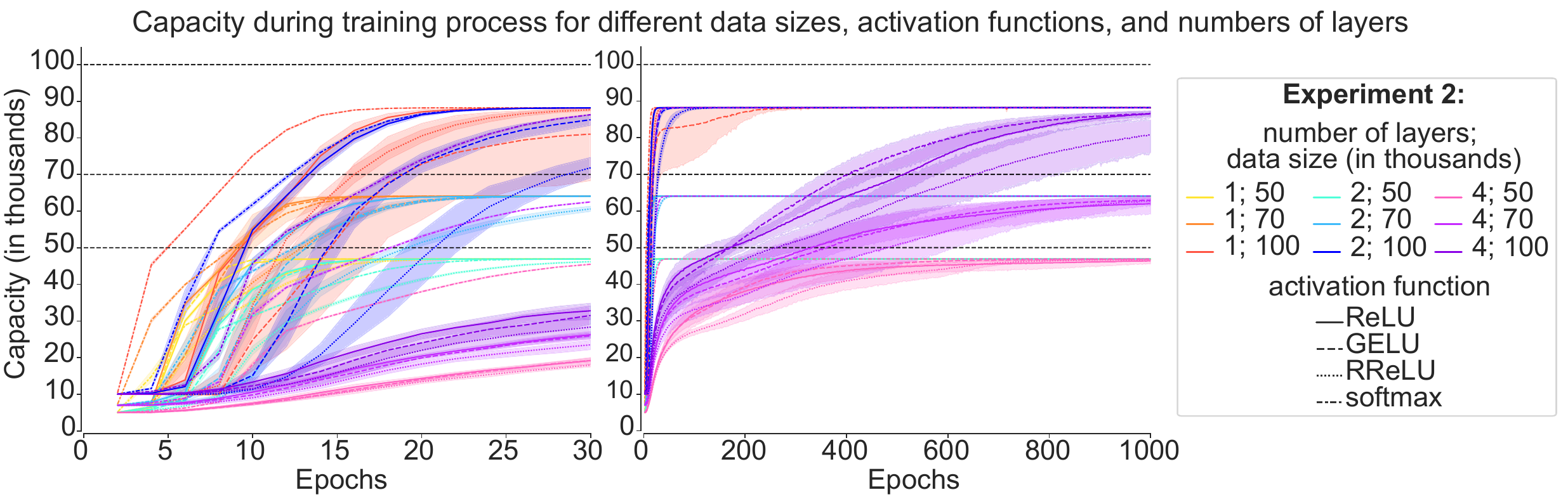}
    \caption{Trends in training capacity for the second setup (different data sizes, activation functions, and numbers of layers for triplets dataset). Left: first $30$ epochs; right: full training process of 1000 epochs.}
    \label{fig3}
\end{figure*}

\begin{figure*}[ht]
    \centering
    \includegraphics[width=0.99\textwidth]{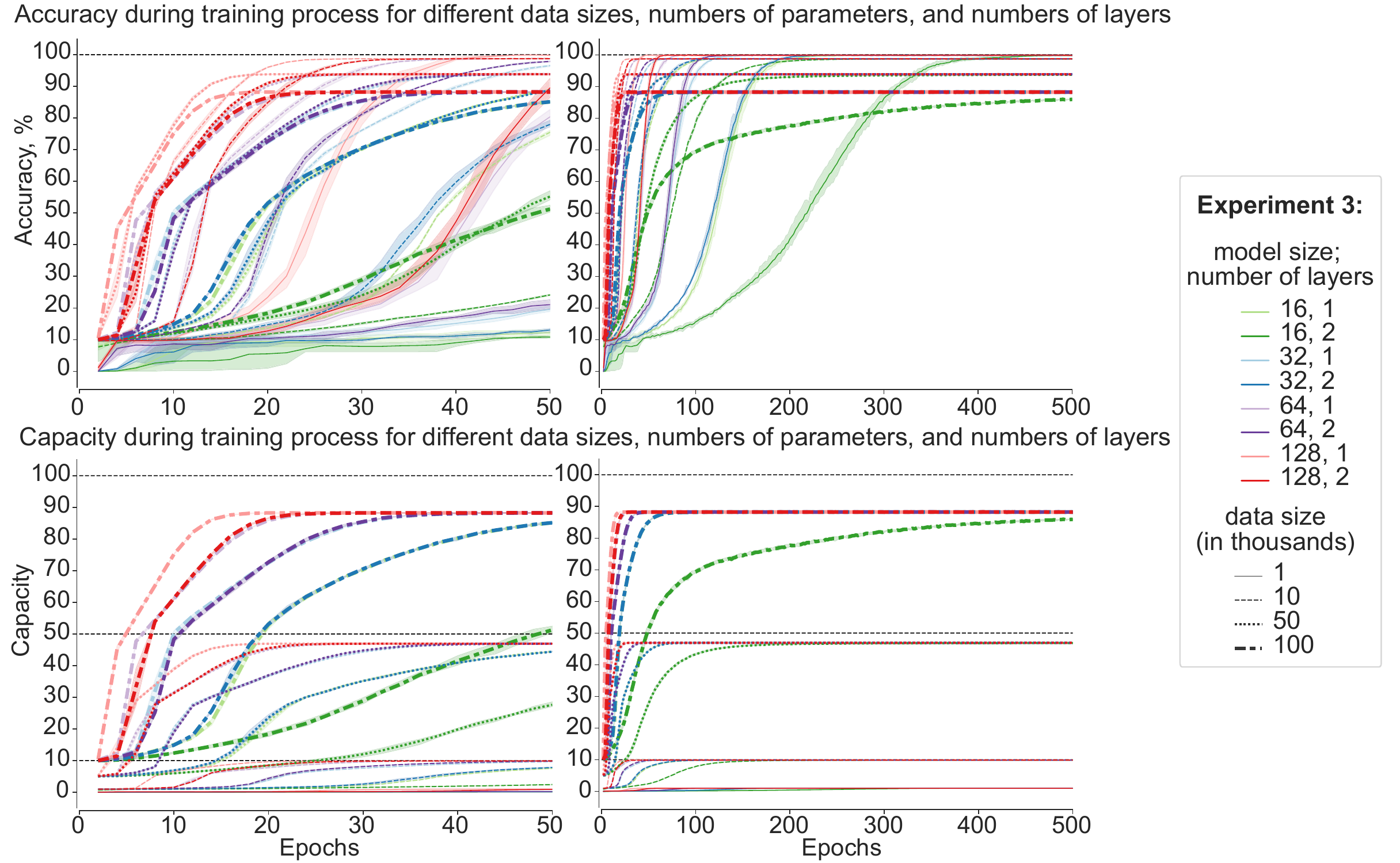}
    \caption{
    Trends in training accuracy (upper) and capacity (lower) for the third setup (different data sizes, numbers of parameters, and numbers of layers for triplets dataset). Left: first $50$ epochs; right: full training process of $500$ epochs. Light color corresponds to $1$ layer, dark – to $2$; number of parameters is a total number for all layers: green – $16$, blue – $32$, violet – $64$, red – $128$; embedding size can be computed by dividing it by layer count. }
    \label{fig4}
\end{figure*}
\begin{figure*}[ht]
    \centering
    \includegraphics[width=0.99\textwidth]{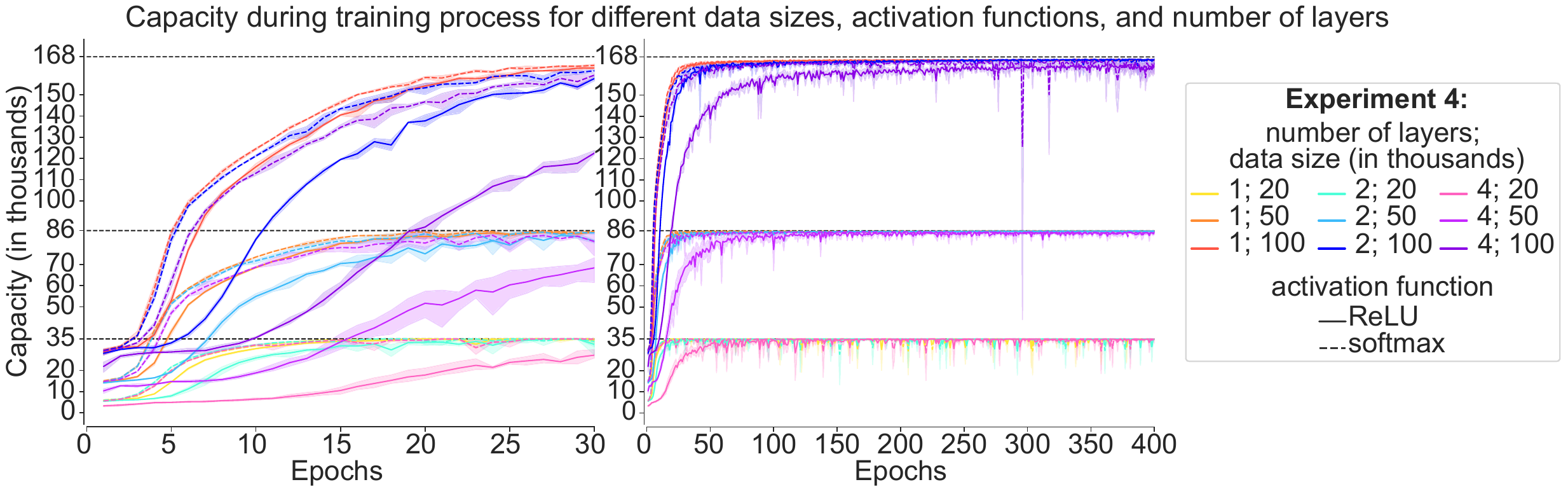}
    \caption{Trends in training capacity for the fourth setup (different data sizes, activation functions, and numbers of layers for sequences dataset). Left: first $30$ epochs; right: full training process of $400$ epochs. }
    \label{fig5}
\end{figure*}

\subsection{Dataset Size Influence}

Figure \ref{fig2} illustrates capacity and accuracy trends across dataset sizes in the first setup. Smaller datasets learn quickly, with both metrics rising rapidly in the first 5–6 epochs and reaching maximum capacity by epoch 20. Larger datasets improve little in the first 15 epochs but later reach higher final accuracy and capacity. This suggests a threshold existence ($\sim70{,}000$ rows for this case), beyond which the training process changes and a lot more epochs are required for full memorization.

The final accuracy and capacity (Table \ref{table1}) indicate that although smaller datasets initially achieve higher accuracy, their capacity remains well below the size of the dataset (e.g., $50{,}000$ rows yield only $46{,}811$ samples). In contrast, larger datasets, such as $100{,}000$ rows, significantly improve memorization ($86{,}776$ samples), highlighting the model's ability to use more data. The progressive increase in capacity suggests that the size of the dataset plays a crucial role in optimizing memorization; however, the reasons behind the unlearned data, despite the available capacity, remain unclear.

\begin{table}[ht]
  \centering
\begin{tabular}{ccc}
\hline
\textbf{data size} & \textbf{accuracy, \%} & \textbf{capacity} \\ \hline
\textbf{$50{,}000$}     & $93.62\pm0.3$         & $46{,}811\pm149$          \\ 
\textbf{$60{,}000$}     & $92.42\pm0.2$          & $55{,}455\pm126$          \\ 
\textbf{$70{,}000$}     & $91.1\pm1.08$          & $63{,}773\pm756$         \\ 
\textbf{$80{,}000$}     & $89.63\pm1.66$         & $71{,}706\pm1326$         \\ 
\textbf{$90{,}000$}     & $87.24\pm1.66$        & $78{,}517\pm2173$        \\ 
\textbf{$100{,}000$}    & $86.78\pm2.42$        & $86{,}776\pm2484$        \\ \hline
\end{tabular}

 \caption{Final results after the full training process for the first setup (data sizes, for triplets dataset).}
  \label{table1}
\end{table}

\subsection{Architectural Variations Influences}

In the second setup, the batch size was increased from $64$ to $128$,  
Since larger batch sizes seem to reduce gradient noise and improve memorization. As a result, one-layer models converged faster and reached higher capacity than in the first setup.

Softmax consistently outperformed other activation functions, yielding the highest average capacity, fewer outliers, and more stable training. Notably, four-layer models with Softmax achieved capacities comparable to one- or two-layer models without sacrificing convergence speed (Figure \ref{fig3}), suggesting its scalability with depth.

In contrast, ReLU and RReLU showed moderate performance, but suffered from increased variability and decreased capacity as the layers increased, aligning with the findings of \citet{paik2023disharmonybnrelucauses} and \citet{chen2024neuralcharacteristicactivationanalysis}. These activations exhibited inconsistent learning patterns, with unexpected slowdowns in capacity improvements \citep{fu2024breakinglearningplateausincontext}. GELU followed a similar trend, though it performed better in the early training stages with larger datasets.

As previously, the size of the dataset significantly affected training: larger sets required longer warm-up phases, initially achieving lower capacities than smaller datasets under the same conditions. This suggests the existence of distinct learning phases where improvements depend on architectural depth, dataset size, and activation function.

Furthermore, adding more layers did not improve performance; instead, it slowed training and reduced final capacity, likely due to the simplicity of the dataset, where additional layers do not provide any advantage in capturing patterns. Although deeper architectures benefit more complex datasets \cite{he2024matterstransformersattentionneeded}, their impact can be reduced for data with simple relationships.

\subsection{Number of Parameters Influence}

The third experiment further confirmed that, for simple datasets, learning dynamics depend on embedding size, not the number of layers. Models with the same embedding size but different layer counts exhibited nearly identical accuracy improvement. For instance, as shown in Figure \ref{fig4}, a one-layer model with $16$ parameters (embedding size is $16$, light green) converged at almost the same rate as a two-layer transformer with $32$ parameters (embedding size is $16$ per layer, dark blue). Similar trends were observed for models with embedding sizes of $32$ and $64$, regardless of layer count.

These results highlight that embedding size is the key factor influencing learning speed, while adding layers without increasing embedding size neither accelerates convergence nor improves final capacity. In fact, additional layers often slow the training, as evidenced by the faster growth of accuracy of one-layer models (Figure \ref{fig4}). Smaller embedding sizes further reduced the learning speed, consistent with previous experiments. However, all configurations ultimately reached similar accuracy, highlighting that the simplicity of the dataset allows embedding size to dominate training dynamics.

The final capacity values remained nearly identical across configurations, regardless of embedding size or layer count: with a dataset size of $1{,}000$ samples, the capacities for the one- and two-layer models were nearly accurate. Similarly, at $10{,}000$ and $50{,}000$ samples, one-layer models achieved $9{,}874\pm11$ and $46{,}939\pm105$, while two-layer models reached $9{,}875\pm7$ and $46{,}911\pm117$, respectively. However, at $100{,}000$ samples, a capacity "barrier" emerged. Two-layer transformers with an embedding size of $8$ ($16$ total parameters) showed the capacity drop to $85{,}935\pm153$, compared to $\sim88{,}200$ for other configurations, while one-layer models maintained a higher capacity of  $88{,}240\pm62$. This suggests that larger datasets, smaller embeddings, and deeper architectures may introduce limitations due to slower convergence or suboptimal capacity utilization.


\subsection{Insights from Sequence Datasets}

In the fourth setup, model capacity was evaluated by testing its ability to memorize each node in a sequence using the full preceding sequence of nodes and edges (instead of triplets), involving $34{,}908$, $85{,}972$, and $167{,}965$ predictions for datasets of $20$, $50$, and $100$ thousand sequences, respectively.

Compared to triplet datasets, models trained on sequences achieved near-perfect memorization in significantly fewer epochs, plateauing within $150$ epochs (Figure \ref{fig5}). The sequential structure likely sped up learning but increased training time because of more information per sequence. Training showed greater capacity fluctuations over epochs, probably reflecting the increased complexity of the dataset, as sequences encode more intricate patterns than triplets. Nonetheless, models demonstrated exceptional memorization, achieving $100\%$ capacity for the $20$ thousand sequence dataset and over $99.5\%$ for $50$ and $100$ thousand sequences.

As before, RReLU converged more slowly than Softmax, however, the final capacities were nearly identical for one- and two-layer models: with $100$ thousand sequences, RReLU achieved $166{,}934\pm243$ (one layer) and $166{,}995\pm118$ (two layers), while Softmax reached $166{,}992\pm110$ and $166{,}985\pm904$, respectively. In deeper models (4 layers), RReLU showed lower final capacities and greater fluctuations ($165{,}271\pm1{,}068$ vs. $166{,}825\pm319$ for Softmax). This contrasts with previous findings \citep{shen2023studyrelusoftmaxtransformer}, which reported that ReLU outperformed Softmax. The discrepancy may suggest that the relative effectiveness of activation functions depends on the dataset structure and task, warranting further investigation. Nonetheless, even with increased sequence complexity, all models demonstrated rapid adaptation and strong memorization.

\section{Discussion}

This study examined how decoder-only transformer models memorize structured data derived from a real-world medical ontology. Our focus was not on generalization, but on a controlled analysis of memorization, presenting a proof-of-concept framework that bridges theoretical insights and practical evaluation. The complete SNOMED KG contains more than a million relations, integrating diverse fields of medicine (e.g., substances, diseases, and anatomical structures). However, in mobile applications, e.g. small transformers in smart glasses or smartwatches, models must efficiently retain only targeted subsets of information. For example, smart glasses for a cardiac surgeon or a smartwatch with a personal dietary coach might require a domain-specific LLM that memorizes about 10 to 100,000 items. As discussed in \citet{kajitsuka2024optimalmemorizationcapacitytransformers, härmä2024empiricalcapacitymodelselfattention}, isolating memorization is a valid objective that reveals how much a transformer can reliably store under different architectural configurations. Our methodology reflects this: we analyze how dataset characteristics and architectural choices affect convergence and memorization, independent of generalization ability or test-time reasoning.

To ensure clear capacity measurement, we deliberately focused on tasks where ground-truth memorization can be unambiguously defined. Increasing complexity would blur the line between memorization and generalization, making interpretation less fair and direct.

\subsection{Effect of Dataset Structure}

Smaller datasets led to faster convergence but lower capacity, whereas larger datasets required longer warm-up but achieved higher memorization. Beyond a certain size, the training slowed significantly, indicating optimization bottlenecks. The fact that some samples remain unlearned even with sufficient capacity points to possible optimization barriers or local minima (see Limitations).

Sequence-based datasets outperformed triplets, achieving near-perfect memorization with fewer epochs. Sequences improved learning by capturing relationships and patterns in the data, though they also led to increased training fluctuations, aligning with \citet{ju2021chunkformerlearninglongtime}. This suggests that longer traversal sequences could further improve memorization in domain-specific medical applications.

The complexity of the sequence datasets was controlled through BFS depth and edge count, allowing capture of both local and global structures from the SNOMED graph (e.g., transitions between anatomical concepts and related procedures), while avoiding trivially linear or purely synthetic patterns. Randomness was balanced with structural constraints such as bidirectional edges and node uniqueness, reflecting how medical knowledge is typically reasoned over in practice (e.g., from symptom to diagnosis to treatment).

\subsection{Architectural Influence}

Embedding size was the main factor in learning speed and capacity, and adding layers often reduced performance, probably due to the data simplicity. This supports the findings that many transformer layers are redundant and can be pruned without loss \citet{he2024matterstransformersattentionneeded}. Although we did not directly analyze redundancy, our results suggest that pruning could further optimize capacity.

For larger datasets, smaller embeddings struggled to reach full capacity, particularly in deeper architectures, suggesting that increasing embedding size is more beneficial than adding depth, at least for structured domain-specific memorization.

Softmax led to greater stability and capacity, while ReLU-based activations showed higher variability and performance drops in deeper models, which is consistent with, e.g., \citet{paik2023disharmonybnrelucauses, chen2024neuralcharacteristicactivationanalysis}. However, this contrasts with \citet{shen2023studyrelusoftmaxtransformer}, who found ReLU advantageous, emphasizing that activation effectiveness may be highly dependent on the structure of the dataset, the initialization of the model, or the formulation of the task.

For deployment in limited edge devices, our results suggest favoring shallow architectures (1 to 2 layers) with wider embeddings, which consistently demonstrated better memorization per parameter. This configuration offers a practical trade-off for applications where total parameter count and energy use are constrained, such as wearables or low-power clinical decision support tools.

\section{Conclusions}

This study investigated how transformer architecture and dataset structure influence memorization capacity, introducing a practical framework for evaluating memorization on real-world data, such as the SNOMED knowledge graph.

Key findings show that embedding size and activation function have more impact than depth, while larger datasets improved memorization but required longer training. Triplets performed well in simpler models, whereas sequences excelled but introduced fluctuations. Challenges remain in efficiency, layer-specific contributions, and generalization, necessitating further research on scalability, compression, and architecture optimization.

For practical use of small transformers in medical smart devices, models must efficiently store specialized knowledge while maintaining computational feasibility. Future work should explore longer sequences, adaptive memory compression, and layer-wise analysis to enhance structured knowledge memorization in practical deployments.

\section{Limitations}

Although this study provides meaningful information, several open questions remain:
\begin{itemize}
    \item Misclassification patterns were not systematically analyzed; unlearned samples may be the result of optimization bottlenecks or data-specific challenges. Strategies, such as curriculum learning \cite{kim2024strategicdataorderingenhancing}, or loss re-weighting \cite{sow2025dynamiclossbasedsamplereweighting} could address these gaps.
    \item  Future research should test these findings on longer sequences and larger datasets to confirm them at scale.
    \item  Layer similarity or redundancy was not directly assessed; future probing and pruning studies (see \citet{allenzhu2024physicslanguagemodels31}) could clarify each layer’s role and enhance efficiency.
    \item Integrating sparse autoencoders \citep{bricken2023monosemanticity} or transcoders \citep{paulo2025transcodersbeatsparseautoencoders} can help distinguish memorization from generalization, clarifying whether certain layers store specific relationships or contribute to greater generalizability.
    \item While proposed sequence generation method reflects realistic ontology traversal, more explicit alignment with clinical reasoning patterns (e.g., decision trees or symptom pathways) is an open direction. Testing on other biomedical graphs, such as GenomicKB \citep{10.1093/nar/gkac957}, which encodes large-scale genomic and transcriptomic relationships, could assess whether memorization patterns generalize to domains with different graph structures.
    
    \item We did not conduct experiments under quantization or activation sparsity constraints, which may affect architectural recommendations for edge applications and warrant follow-up work.

\end{itemize}

Addressing these limitations will further refine transformer optimization strategies for structured data modeling and knowledge retention.

\section{Acknowledgments}
We acknowledge SURF for providing computational resources via Snellius, the Dutch National Supercomputer, which facilitated large-scale model training \citep{surf2024snellius}.



\begin{thebibliography}{31}
\providecommand{\natexlab}[1]{#1}

\bibitem[{Agarap(2019)}]{agarap2019deeplearningusingrectified}
Abien~Fred Agarap. 2019.
\newblock \href {https://arxiv.org/abs/1803.08375} {Deep learning using rectified linear units (relu)}.
\newblock \emph{Preprint}, arXiv:1803.08375.

\bibitem[{Allen-Zhu and Li(2024)}]{allenzhu2024physicslanguagemodels31}
Zeyuan Allen-Zhu and Yuanzhi Li. 2024.
\newblock \href {https://arxiv.org/abs/2309.14316} {Physics of language models: Part 3.1, knowledge storage and extraction}.
\newblock \emph{Preprint}, arXiv:2309.14316.

\bibitem[{Balloccu et~al.(2024)Balloccu, Reiter, Kumar, Recupero, and Riboni}]{balloccu_ask_2024}
Simone Balloccu, Ehud Reiter, Vivek Kumar, Diego~Reforgiato Recupero, and Daniele Riboni. 2024.
\newblock \href {https://doi.org/10.48550/arXiv.2401.08420} {Ask the experts: sourcing high-quality datasets for nutritional counselling through {Human}-{AI} collaboration}.
\newblock \emph{arXiv preprint}.
\newblock ArXiv:2401.08420 [cs].

\bibitem[{Boltzmann(1868)}]{boltzmann1868studien}
Ludwig Boltzmann. 1868.
\newblock Studien über das gleichgewicht der lebendigen kraft zwischen bewegten materiellen punkten.
\newblock \emph{Wiener Berichte}, 58:517--560.
\newblock Studies on the balance of living force between moving material points.

\bibitem[{Bricken et~al.(2023)Bricken, Templeton, Batson, Chen, Jermyn, Conerly, Turner, Anil, Denison, Askell, Lasenby, Wu, Kravec, Schiefer, Maxwell, Joseph, Hatfield-Dodds, Tamkin, Nguyen, McLean, Burke, Hume, Carter, Henighan, and Olah}]{bricken2023monosemanticity}
Trenton Bricken, Adly Templeton, Joshua Batson, Brian Chen, Adam Jermyn, Tom Conerly, Nicholas~L. Turner, Cem Anil, Carson Denison, Amanda Askell, Robert Lasenby, Yifan Wu, Shauna Kravec, Nicholas Schiefer, Tim Maxwell, Nicholas Joseph, Zach Hatfield-Dodds, Alex Tamkin, Karina Nguyen, Brayden McLean, Josiah~E. Burke, Tristan Hume, Shan Carter, Tom Henighan, and Chris Olah. 2023.
\newblock \href {https://transformer-circuits.pub/2023/monosemantic-features} {Towards monosemanticity: Decomposing language models with dictionary learning}.
\newblock \emph{Transformer Circuits Thread}.

\bibitem[{Brown et~al.(2020)Brown, Mann, Ryder, Subbiah, Kaplan, Dhariwal, Neelakantan, Shyam, Sastry, Askell, Agarwal, Herbert-Voss, Krueger, Henighan, Child, Ramesh, Ziegler, Wu, Winter, Hesse, Chen, Sigler, Litwin, Gray, Chess, Clark, Berner, McCandlish, Radford, Sutskever, and Amodei}]{brown2020languagemodelsfewshotlearners}
Tom~B. Brown, Benjamin Mann, Nick Ryder, Melanie Subbiah, Jared Kaplan, Prafulla Dhariwal, Arvind Neelakantan, Pranav Shyam, Girish Sastry, Amanda Askell, Sandhini Agarwal, Ariel Herbert-Voss, Gretchen Krueger, Tom Henighan, Rewon Child, Aditya Ramesh, Daniel~M. Ziegler, Jeffrey Wu, Clemens Winter, Christopher Hesse, Mark Chen, Eric Sigler, Mateusz Litwin, Scott Gray, Benjamin Chess, Jack Clark, Christopher Berner, Sam McCandlish, Alec Radford, Ilya Sutskever, and Dario Amodei. 2020.
\newblock \href {https://arxiv.org/abs/2005.14165} {Language models are few-shot learners}.
\newblock \emph{Preprint}, arXiv:2005.14165.

\bibitem[{Chen and Ge(2024)}]{chen2024neuralcharacteristicactivationanalysis}
Wenlin Chen and Hong Ge. 2024.
\newblock \href {https://arxiv.org/abs/2305.15912} {Neural characteristic activation analysis and geometric parameterization for relu networks}.
\newblock \emph{Preprint}, arXiv:2305.15912.

\bibitem[{El-Sappagh et~al.(2018)El-Sappagh, Franda, Ali, and Kwak}]{El-Sappagh2018}
Shaker El-Sappagh, Francesco Franda, Farman Ali, and Kyung-Sup Kwak. 2018.
\newblock \href {https://doi.org/10.1186/s12911-018-0651-5} {Snomed ct standard ontology based on the ontology for general medical science}.
\newblock \emph{BMC Medical Informatics and Decision Making}, 18(1):76.

\bibitem[{Feng et~al.(2022)Feng, Tang, Gao, Zhu, Li, Yang, Yao, Huang, and Liu}]{10.1093/nar/gkac957}
Fan Feng, Feitong Tang, Yijia Gao, Dongyu Zhu, Tianjun Li, Shuyuan Yang, Yuan Yao, Yuanhao Huang, and Jie Liu. 2022.
\newblock \href {https://doi.org/10.1093/nar/gkac957} {Genomickb: a knowledge graph for the human genome}.
\newblock \emph{Nucleic Acids Research}, 51(D1):D950--D956.

\bibitem[{Fu et~al.(2024)Fu, Yang, Wang, Lu, and Zheng}]{fu2024breakinglearningplateausincontext}
Jingwen Fu, Tao Yang, Yuwang Wang, Yan Lu, and Nanning Zheng. 2024.
\newblock \href {https://arxiv.org/abs/2309.06054} {Breaking through the learning plateaus of in-context learning in transformer}.
\newblock \emph{Preprint}, arXiv:2309.06054.

\bibitem[{Gupta et~al.(2024)Gupta, Ta, Ram, and Sivaprakasam}]{gupta_comprehensive_2024}
Bhumika Gupta, Pralaypati Ta, Keerthi Ram, and Mohanasankar Sivaprakasam. 2024.
\newblock \href {https://doi.org/10.1109/CIBCB58642.2024.10702112} {Comprehensive {Modeling} and {Question} {Answering} of {Cancer} {Clinical} {Practice} {Guidelines} using {LLMs}}.
\newblock In \emph{2024 {IEEE} {Conference} on {Computational} {Intelligence} in {Bioinformatics} and {Computational} {Biology} ({CIBCB})}, pages 1--8.
\newblock ISSN: 2994-9408.

\bibitem[{He et~al.(2024)He, Sun, Shen, and Li}]{he2024matterstransformersattentionneeded}
Shwai He, Guoheng Sun, Zheyu Shen, and Ang Li. 2024.
\newblock \href {https://arxiv.org/abs/2406.15786} {What matters in transformers? not all attention is needed}.
\newblock \emph{Preprint}, arXiv:2406.15786.

\bibitem[{Hendrycks and Gimpel(2023)}]{hendrycks2023gaussianerrorlinearunits}
Dan Hendrycks and Kevin Gimpel. 2023.
\newblock \href {https://arxiv.org/abs/1606.08415} {Gaussian error linear units (gelus)}.
\newblock \emph{Preprint}, arXiv:1606.08415.

\bibitem[{Härmä et~al.(2024)Härmä, Pietrasik, and Wilbik}]{härmä2024empiricalcapacitymodelselfattention}
Aki Härmä, Marcin Pietrasik, and Anna Wilbik. 2024.
\newblock \href {https://arxiv.org/abs/2407.15425} {Empirical capacity model for self-attention neural networks}.
\newblock \emph{Preprint}, arXiv:2407.15425.

\bibitem[{Ju et~al.(2021)Ju, Isac, and Nie}]{ju2021chunkformerlearninglongtime}
Yue Ju, Alka Isac, and Yimin Nie. 2021.
\newblock \href {https://arxiv.org/abs/2112.15087} {Chunkformer: Learning long time series with multi-stage chunked transformer}.
\newblock \emph{Preprint}, arXiv:2112.15087.

\bibitem[{Kajitsuka and Sato(2024)}]{kajitsuka2024optimalmemorizationcapacitytransformers}
Tokio Kajitsuka and Issei Sato. 2024.
\newblock \href {https://arxiv.org/abs/2409.17677} {Optimal memorization capacity of transformers}.
\newblock \emph{Preprint}, arXiv:2409.17677.

\bibitem[{Kim and Lee(2024)}]{kim2024strategicdataorderingenhancing}
Jisu Kim and Juhwan Lee. 2024.
\newblock \href {https://arxiv.org/abs/2405.07490} {Strategic data ordering: Enhancing large language model performance through curriculum learning}.
\newblock \emph{Preprint}, arXiv:2405.07490.

\bibitem[{Kim et~al.(2023)Kim, Kim, and Mozafari}]{kim2023provable}
Junghwan Kim, Michelle Kim, and Barzan Mozafari. 2023.
\newblock \href {https://openreview.net/forum?id=8JCg5xJCTPR} {Provable memorization capacity of transformers}.
\newblock In \emph{The Eleventh International Conference on Learning Representations}.

\bibitem[{Kingma and Ba(2017)}]{kingma2017adammethodstochasticoptimization}
Diederik~P. Kingma and Jimmy Ba. 2017.
\newblock \href {https://arxiv.org/abs/1412.6980} {Adam: A method for stochastic optimization}.
\newblock \emph{Preprint}, arXiv:1412.6980.

\bibitem[{Lamy(2017)}]{lamy2017owlready}
Jean-Baptiste Lamy. 2017.
\newblock \href {https://doi.org/10.1016/j.artmed.2017.07.002} {Owlready: Ontology-oriented programming in python with automatic classification and high level constructs for biomedical ontologies}.
\newblock \emph{Artificial Intelligence in Medicine}, 80:11--28.

\bibitem[{Mahdavi et~al.(2024)Mahdavi, Liao, and Thrampoulidis}]{mahdavi2024memorizationcapacitymultiheadattention}
Sadegh Mahdavi, Renjie Liao, and Christos Thrampoulidis. 2024.
\newblock \href {https://arxiv.org/abs/2306.02010} {Memorization capacity of multi-head attention in transformers}.
\newblock \emph{Preprint}, arXiv:2306.02010.

\bibitem[{Paik and Choi(2023)}]{paik2023disharmonybnrelucauses}
Inyoung Paik and Jaesik Choi. 2023.
\newblock \href {https://arxiv.org/abs/2304.11692} {The disharmony between bn and relu causes gradient explosion, but is offset by the correlation between activations}.
\newblock \emph{Preprint}, arXiv:2304.11692.

\bibitem[{Paszke et~al.(2017)Paszke, Gross, Chintala, Chanan, Yang, DeVito, Lin, Desmaison, Antiga, and Lerer}]{paszke2017automatic}
Adam Paszke, Sam Gross, Soumith Chintala, Gregory Chanan, Edward Yang, Zachary DeVito, Zeming Lin, Alban Desmaison, Luca Antiga, and Adam Lerer. 2017.
\newblock Automatic differentiation in pytorch.

\bibitem[{Paulo et~al.(2025)Paulo, Shabalin, and Belrose}]{paulo2025transcodersbeatsparseautoencoders}
Gonçalo Paulo, Stepan Shabalin, and Nora Belrose. 2025.
\newblock \href {https://arxiv.org/abs/2501.18823} {Transcoders beat sparse autoencoders for interpretability}.
\newblock \emph{Preprint}, arXiv:2501.18823.

\bibitem[{Shehzad et~al.(2024)Shehzad, Xia, Abid, Peng, Yu, Zhang, and Verspoor}]{shehzad2024graphtransformerssurvey}
Ahsan Shehzad, Feng Xia, Shagufta Abid, Ciyuan Peng, Shuo Yu, Dongyu Zhang, and Karin Verspoor. 2024.
\newblock \href {https://arxiv.org/abs/2407.09777} {Graph transformers: A survey}.
\newblock \emph{Preprint}, arXiv:2407.09777.

\bibitem[{Shen et~al.(2023)Shen, Guo, Tan, Tang, Wang, and Bian}]{shen2023studyrelusoftmaxtransformer}
Kai Shen, Junliang Guo, Xu~Tan, Siliang Tang, Rui Wang, and Jiang Bian. 2023.
\newblock \href {https://arxiv.org/abs/2302.06461} {A study on relu and softmax in transformer}.
\newblock \emph{Preprint}, arXiv:2302.06461.

\bibitem[{Sow et~al.(2025)Sow, Woisetschläger, Bulusu, Wang, Jacobsen, and Liang}]{sow2025dynamiclossbasedsamplereweighting}
Daouda Sow, Herbert Woisetschläger, Saikiran Bulusu, Shiqiang Wang, Hans-Arno Jacobsen, and Yingbin Liang. 2025.
\newblock \href {https://arxiv.org/abs/2502.06733} {Dynamic loss-based sample reweighting for improved large language model pretraining}.
\newblock \emph{Preprint}, arXiv:2502.06733.

\bibitem[{{SURF}(2024)}]{surf2024snellius}
{SURF}. 2024.
\newblock \href {https://servicedesk.surf.nl/wiki/display/WIKI/Snellius} {Snellius - the dutch national supercomputer}.

\bibitem[{Wolf et~al.(2019)Wolf, Debut, Sanh, Chaumond, Delangue, Moi, Cistac, Rault, Louf, Funtowicz, and Brew}]{DBLP:journals/corr/abs-1910-03771}
Thomas Wolf, Lysandre Debut, Victor Sanh, Julien Chaumond, Clement Delangue, Anthony Moi, Pierric Cistac, Tim Rault, R{\'{e}}mi Louf, Morgan Funtowicz, and Jamie Brew. 2019.
\newblock \href {https://arxiv.org/abs/1910.03771} {Huggingface's transformers: State-of-the-art natural language processing}.
\newblock \emph{CoRR}, abs/1910.03771.

\bibitem[{Wu et~al.(2024)Wu, Liang, Bai, and Chen}]{wu_surgbox_2024}
Jinlin Wu, Xusheng Liang, Xuexue Bai, and Zhen Chen. 2024.
\newblock \href {https://doi.org/10.48550/arXiv.2412.05187} {{SurgBox}: {Agent}-{Driven} {Operating} {Room} {Sandbox} with {Surgery} {Copilot}}.
\newblock \emph{arXiv preprint}.
\newblock ArXiv:2412.05187 [cs].

\bibitem[{Xu et~al.(2015)Xu, Wang, Chen, and Li}]{xu2015empiricalevaluationrectifiedactivations}
Bing Xu, Naiyan Wang, Tianqi Chen, and Mu~Li. 2015.
\newblock \href {https://arxiv.org/abs/1505.00853} {Empirical evaluation of rectified activations in convolutional network}.
\newblock \emph{Preprint}, arXiv:1505.00853.

\end{thebibliography}

\appendix

\newpage

\section{Appendix: Additional Representations of the Results}
\label{sec:appendix}

\begin{figure*}[htb!]
    \centering
    \includegraphics[width=\textwidth]{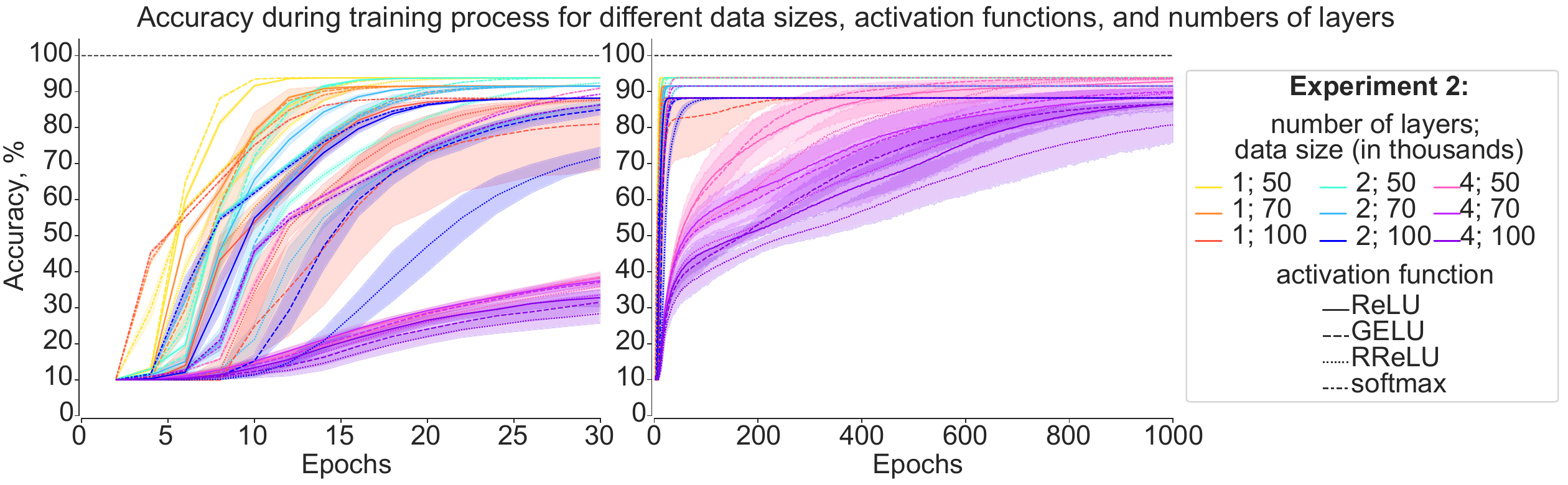}
    \caption{Trends in training accuracy for the second setup (different data sizes, activation functions, and numbers of layers for triplets dataset). Left: first $30$ epochs; right: full training process of $1000$ epochs.}
    \label{SF1}
\end{figure*}

\begin{figure*}[ht]
    \centering
    \includegraphics[width=\textwidth]{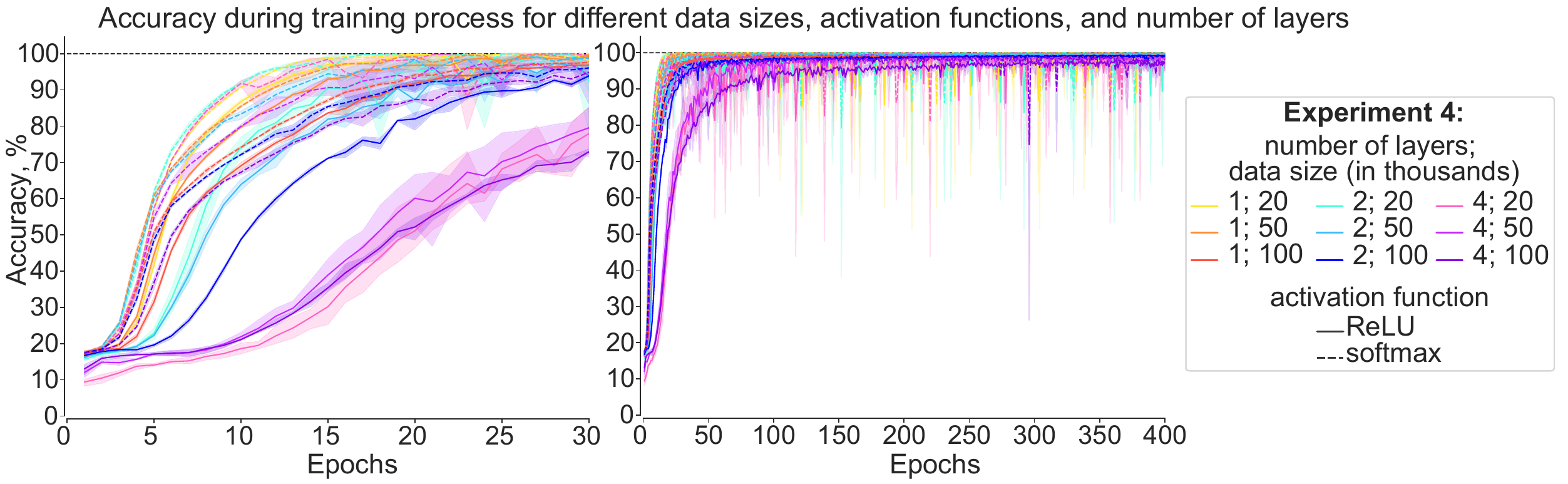}
    \caption{Trends in training accuracy for the fourth setup (different data sizes, activation functions, and numbers of layers for sequences dataset). Left: first $30$ epochs; right: full training process of $400$ epochs. }
    \label{SF2}
\end{figure*}

This appendix provides supplementary visualizations and tables for the experiments conducted:

\begin{itemize}
    \item Second experiment:
    \begin{itemize}
        \item Figure \ref{SF1}: Accuracy trends during training.
        \item Table \ref{ST1}: Final capacities.
    \end{itemize}
    \item Third experiment:
    \begin{itemize}
        \item Table \ref{ST2}: Final capacities.
    \end{itemize}
    \item Fourth experiment:
    \begin{itemize}
        \item Figure \ref{SF2}: Accuracy trends during training.
        \item Table \ref{ST3}: Final capacities.
    \end{itemize}
\end{itemize}

\begin{table*}[ht]
  \centering
\begin{tabular}{c|c|ccc}
\hline
\multirow{2}{*}{\textbf{activation function}}   & \multirow{2}{*}{\textbf{layers count}}  & \multicolumn{3}{c}{\textbf{data sizes}}                                       \\ \cline{3-5} 
      && \multicolumn{1}{c}{\textbf{$50{,}000$}} & \multicolumn{1}{c}{\textbf{$70{,}000$}} & \textbf{$100{,}000$} \\ \hline
\multirow{3}{*}{\textbf{ReLU}}  & \textbf{$1$} & \multicolumn{1}{c}{$46{,}898\pm158$}   & \multicolumn{1}{c}{$64{,}091\pm192$}   & $88{,}148\pm312$  \\ 
                                & \textbf{$2$} & \multicolumn{1}{c}{$46{,}920\pm112$}   & \multicolumn{1}{c}{$64{,}086\pm130$}   & $88{,}217\pm125$   \\ 
                                & \textbf{$4$} & \multicolumn{1}{c}{$46{,}391\pm2{,}268$} & \multicolumn{1}{c}{$61{,}931\pm8{,}480$} & $86{,}558\pm3{,}291$ \\ \hline
\multirow{3}{*}{\textbf{GELU}}  & \textbf{$1$} & \multicolumn{1}{c}{$46{,}925\pm105$}   & \multicolumn{1}{c}{$64{,}096\pm184$}   & $88{,}195\pm123$   \\ 
                                & \textbf{$2$} & \multicolumn{1}{c}{$46{,}926\pm115$}   & \multicolumn{1}{c}{$64{,}080\pm120$}   & $88{,}215\pm128$   \\ 
                                & \textbf{$4$} & \multicolumn{1}{c}{$46{,}798\pm156$}   & \multicolumn{1}{c}{$62{,}949\pm1{,}906$}  & $86{,}589\pm2{,}202$ \\ \hline
\multirow{3}{*}{\textbf{RReLU}} & \textbf{$1$} & \multicolumn{1}{c}{$46{,}930\pm125$}   & \multicolumn{1}{c}{$64{,}080\pm122$}   & $88{,}180\pm180$   \\ 
                                & \textbf{$2$} & \multicolumn{1}{c}{$46{,}927\pm121$}   & \multicolumn{1}{c}{$64{,}088\pm117$}   & $88{,}208\pm132$   \\ 
                                & \textbf{$4$} & \multicolumn{1}{c}{$46{,}730\pm223$}  & \multicolumn{1}{c}{$62{,}818\pm3{,}680$} & $80{,}755\pm15{,}844$ \\ \hline
\multirow{3}{*}{\textbf{softmax}} & \textbf{$1$}     & \multicolumn{1}{c}{$46{,}924\pm87$}       & \multicolumn{1}{c}{$64{,}082\pm166$}       & $88{,}211\pm192$        \\ 
                                & \textbf{$2$} & \multicolumn{1}{c}{$46{,}908\pm127$}   & \multicolumn{1}{c}{$64{,}074\pm134$}   & $88{,}213\pm171$   \\ 
                                & \textbf{$4$} & \multicolumn{1}{c}{$46{,}923\pm104$}   & \multicolumn{1}{c}{$64{,}085\pm131$}   & $88{,}197\pm134$   \\ \hline
\multirow{3}{*}{\textbf{all}}   & \textbf{$1$} & \multicolumn{1}{c}{$46{,}919\pm119$}   & \multicolumn{1}{c}{$64{,}087\pm162$}   & $88{,}183\pm210$  \\ 
                                & \textbf{$2$} & \multicolumn{1}{c}{$46{,}920\pm115$}   & \multicolumn{1}{c}{$64{,}082\pm121$}   & $88{,}213\pm135$   \\ 
                                & \textbf{$4$} & \multicolumn{1}{c}{$46{,}710\pm1169$}  & \multicolumn{1}{c}{$62{,}945\pm4{,}92$} & $85{,}525\pm9{,}720$ \\ \hline
\end{tabular}

 \caption{Final capacity after the full training process for the second setup (different numbers of layers, data sizes, and activation functions for triplets dataset).}
  \label{ST1}
\end{table*}

\begin{table*}[hbt!]
  \centering
\begin{tabular}{c|c|cccc}
\hline
\multirow{2}{*}{\textbf{embedding parameters}}                              &    \multirow{2}{*}{\textbf{layers count}}                 & \multicolumn{4}{c}{\textbf{data sizes}}                               \\ \cline{3-6}
              &  & \textbf{$1{,}000$} & \textbf{$10{,}000$} & \textbf{$50{,}000$} & \textbf{$100{,}000$} \\ \hline
\multirow{2}{*}{\textbf{16}}  & \textbf{1}          & $ 1{,}000 \pm 1     $&$ 9{,}870 \pm 10     $&$ 46{,}937 \pm 148   $&$ 88{,}236 \pm 74    $ \\
                              & \textbf{2}          &$ 998 \pm 3      $&$ 9{,}875 \pm 4      $&$ 46{,}858 \pm 93    $&$ 85{,}935 \pm 153   $ \\ \hline
\multirow{2}{*}{\textbf{32}}  & \textbf{1}          &$ 998 \pm 3      $&$ 9{,}872 \pm 11     $&$ 46{,}955 \pm 119   $&$ 88{,}234 \pm 62    $ \\
                              & \textbf{2}          &$ 999 \pm 3      $&$ 9{,}876 \pm 9      $&$ 46{,}927 \pm 128   $&$ 88{,}252 \pm 82    $ \\ \hline
\multirow{2}{*}{\textbf{64}}  & \textbf{1}          &$ 999 \pm 2      $&$ 9{,}878 \pm 9      $&$ 46{,}932 \pm 122   $&$ 88{,}242 \pm 102   $ \\
                              & \textbf{2}          &$ 999 \pm 3      $&$ 9{,}876 \pm 7      $&$ 46{,}919 \pm 96    $&$ 88{,}237 \pm 58    $ \\ \hline
\multirow{2}{*}{\textbf{128}} & \textbf{1}          &$ 999 \pm 2      $&$ 9{,}877 \pm 12     $&$ 46{,}930 \pm 85    $&$ 88{,}248 \pm 29    $ \\
                              & \textbf{2}          &$ 999 \pm 3      $&$ 9{,}872 \pm 6      $&$ 46{,}938 \pm 131   $&$ 88{,}214 \pm 53    $ \\ \hline
\multirow{2}{*}{\textbf{all}} & \textbf{1}          &$ 999 \pm 2      $&$ 9{,}874 \pm 11     $&$ 46{,}939 \pm 105   $&$ 88{,}240 \pm 62    $ \\
                              & \textbf{2}          &$ 999 \pm 3      $&$ 9{,}875 \pm 7      $&$ 46{,}911 \pm 117   $&$ 87{,}660 \pm 2{,}082  $ \\ \hline
\end{tabular}

 \caption{Final capacity after the full training process for the third setup (different data sizes, numbers of parameters, and numbers of layers for triplets dataset).}
  \label{ST2}
\end{table*}

\begin{table*}[hbt!]
  \centering
\begin{tabular}{c|c|ccc}
\hline
\multirow{2}{*}{\textbf{activation function}} & \multirow{2}{*}{\textbf{layers count}} & \multicolumn{3}{c}{\textbf{\# of sequences (\# of predictions)}}           \\ \cline{3-5} 
                                              &                                         & \textbf{$20{,}000$ ($34{,}908$)} & \textbf{$50{,}000$ ($85{,}972$)} & \textbf{$100{,}000$ ($167{,}965$)} \\ \hline
\multirow{3}{*}{\textbf{RReLU}}               & \textbf{1}                              &$ 34{,}908 \pm  0    $&$ 85{,}936 \pm 31   $&$ 166{,}934 \pm 243  $\\
                                              & \textbf{2}                              &$ 34{,}908 \pm   0   $&$ 85{,}917 \pm 34   $&$ 166{,}995 \pm 118  $\\
                                              & \textbf{4}                              &$ 34{,}908 \pm   0   $&$ 85{,}647 \pm 270  $&$ 165{,}271 \pm 1{,}068 $\\ \hline
\multirow{3}{*}{\textbf{softmax}}             & \textbf{1}                              &$ 34{,}908 \pm   0   $&$ 85{,}931 \pm 18   $&$ 166{,}992 \pm 110  $\\
                                              & \textbf{2}                              &$ 34{,}908 \pm   0   $&$ 85{,}888 \pm 33   $&$ 166{,}985 \pm 904  $\\
                                              & \textbf{4}                              &$ 34{,}908 \pm   0   $&$ 85{,}771 \pm 42   $&$ 166{,}825 \pm 319 $\\ \hline
\multirow{3}{*}{\textbf{all}}                 & \textbf{1}                              &$ 34{,}908 \pm   0   $&$ 85{,}934 \pm 23   $&$ 166{,}963 \pm 180  $\\
                                              & \textbf{2}                              &$ 34{,}908 \pm   0   $&$ 85{,}903 \pm 44   $&$ 166{,}990 \pm 577  $\\
                                              & \textbf{4}                              &$ 34{,}908 \pm   0   $&$ 85{,}709 \pm 220  $&$ 166{,}048 \pm 1{,}842 $\\ \hline
\end{tabular}

 \caption{Final capacity after the full training process for the fourth setup (different data sizes, activation functions, and numbers of layers for sequences dataset).}
  \label{ST3}
\end{table*}

\end{document}